\PassOptionsToPackage{dvipsnames,table}{xcolor}
\documentclass[11pt]{article}

\usepackage[preprint]{acl}

\usepackage{times}
\usepackage{latexsym}
\usepackage{comment,booktabs}
\usepackage{xcolor}
\usepackage{tikz}
\usepackage{forest}
\usetikzlibrary{arrows.meta}
\usetikzlibrary{shadows}
\usepackage[most]{tcolorbox}
\usepackage{enumitem}
\usepackage{pifont}   % for \ding symbols (optional)
\usepackage{xcolor}

\usepackage[T1]{fontenc}
\usepackage[utf8]{inputenc}

\usepackage{microtype}

\usepackage{inconsolata}

\usepackage{graphicx}

\usepackage{tcolorbox}
\usepackage{amssymb}
\usepackage{enumitem}
\tcbuselibrary{skins,breakable}
\definecolor{plasma1}{HTML}{0D0887}  % Deep purple
\definecolor{plasma2}{HTML}{7E03A8}  % Purple
\definecolor{plasma3}{HTML}{CC4778}  % Magenta-pink
\definecolor{plasma4}{HTML}{F89540}  % Orange
\title{On Challenges in Evaluation Methodology: A Review}

\author{Ruchira Dhar$^{\dagger}$ \ \ 
        Qiwei Peng$^{\dagger}$ \ \ 
        \textbf{Anders S{\o}gaard}$^{\dagger}$ \\
        $^{\dagger}$Department of Computer Science, University of Copenhagen \\ 
        }

\begin{document}
\maketitle

\begin{abstract}
Recent advances in language models have prompted a growing body of work that questions prevailing evaluation practices. However, many such critiques have already been discussed in natural language processing (NLP), a field with a long history of methodological reflection on evaluation. We conduct a scoping review spanning four decades and develop a taxonomy of 22 recurring concerns. For each concern, we examine how recently it was discussed in the reviewed literature and whether it receives explicit attention in institutional checklists, finding that some concerns have become less visible in literature and many lack coverage in these checklists. Our work provides an overview of the major concerns in evaluation methodology that have shaped the field and identifies potential targets for renewed investigation and field-level intervention by the research community.
\end{abstract}

\section{Introduction}

The Natural Language Processing (NLP) community and literature have grown rapidly in recent years, e.g., with the introduction of Large Language Models (LLMs) \cite{radford2019language, brown2020language} with many papers devoted to methodological concerns around evaluation \cite{deshpande2025multichallenge, kazemi2025big, wang2025exploring}: what evaluation results actually indicate \cite{11002710, wallach2025position, wei2025position, salaudeen2025measurement}, what claims they support about model capabilities \cite{10.1145/3708359.3712091, zhou2025lost, singh2025leaderboard}, and how reliably can they be interpreted and used \cite{eriksson2025ai, weidinger2025toward, rane2025position, zhuang2025position, li-etal-2025-exploring-reliability, ye2025justice, chehbouni2025neither}. However, many of these issues echo earlier critiques from NLP. %that have been discussed and debated long in the field of NLP and the rapid growth of NLP risks old discussions keep being forgotten and rediscovered. 
So, to what extent are we reinventing wheels? 

\begin{figure}[t]
    \centering
    \includegraphics[width=\columnwidth]{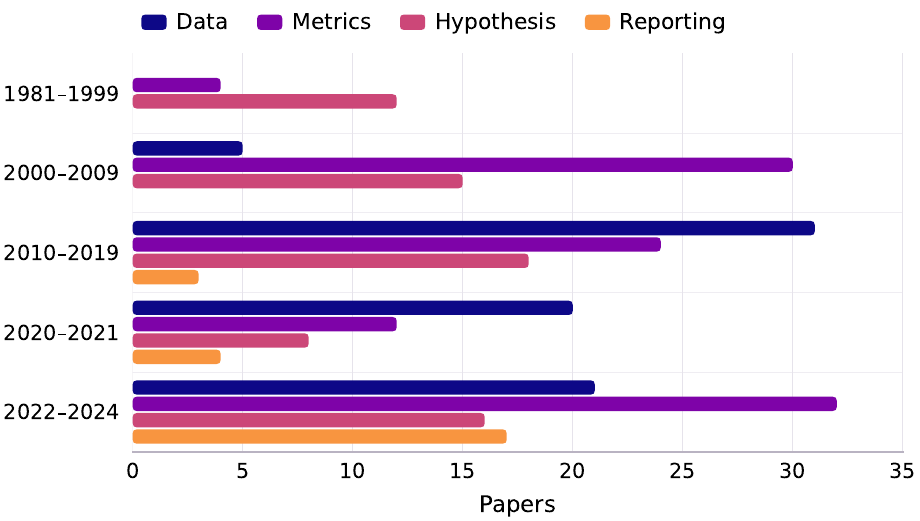}
    \caption{Temporal distribution of survey papers addressing evaluation methodology (1981--2024) across the four main concern types.}
    \label{fig:acl-papers}
\end{figure}

With a long history of grappling with evaluation, NLP has visited and revisited for decades many methodological debates on evaluation \cite{1457848, cohen1988evaluation, jones1994towards, church2019survey}. %This historical continuity suggests that evaluation challenges today are not solely a consequence of recent model scale but reflect deeper methodological tensions in how evaluation is practiced.  Given this, it seems fruitful to take stock of evaluation concerns as they have developed across NLP: \textit{first}, to avoid researchers from reinventing the wheel and rediscovering solutions to problems the field has already worked through; and \textit{second}, to create a consolidated reference that new evaluators can draw on when building benchmarks and designing evaluations. 
Taking stock of the current status of where each evaluation concern stands is thus useful in two ways: \textit{first}, to surface what the field has worked through so researchers are not reimporting settled questions as open ones; and \textit{second}, to identify which concerns remain unresolved and warrant renewed attention from the community. This work is our attempt to do the above by bringing history into conversation with the present\footnote{See \autoref{sec:apprelatedwork} for related work and how our contribution differs from existing work.}, and, offer the field a resource that is both a record of where evaluation has been and a guide for future efforts. 

\paragraph{\textbf{Contributions}}
\textbf{(i)} A review of research in methodological concerns in NLP evaluation;
\textbf{(ii)} a taxonomy organizing these concerns into four dimensions and 22 leaf categories, with discussions on each concern;
\textbf{(iii)} an analysis of recent attention to each concern and its coverage in institutional research checklists, identifying targets for field-level intervention and community action.

%\paragraph{\textbf{Contributions}}\textbf{(i)} A survey of evaluation concerns in NLP  from 1981--2024; \textbf{(ii)} a taxonomy organizing these concerns into four categories and 22 leaf categories; \textbf{(iii)} a classification of each concern by current status, identifying targets for field-level intervention and community action.

\section{Methodological Framework}
\label{sec2:surveymethod}

\begin{figure}[t]
    \centering
    \includegraphics[width=\columnwidth]
    {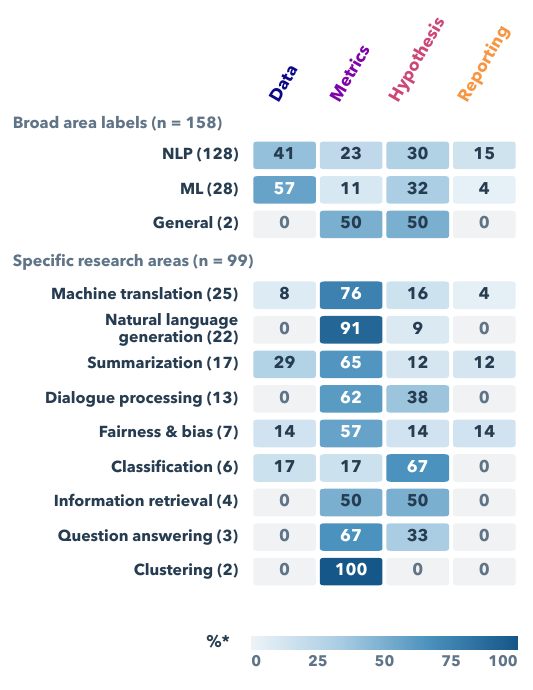}
    \caption{Evaluation concerns across research areas showing the percentage of papers within each area assigned to each taxonomy dimension.}
    \label{fig:area-papers}
\end{figure}

\paragraph{Literature Review}  Given the breadth and diversity of the NLP evaluation literature, we adopt a scoping review \cite{pham2014scoping, peters2021scoping} to synthesize key contributions and identify recurring concerns, rather than provide exhaustive coverage. We derive a final corpus of 257 papers (1981 -- 2024): 192 papers from the ACL Anthology and 65 from the broader ML venues\footnote{Full details of the search, screening, and inclusion criteria are provided in\autoref{sec:appsurveymethod}, with the PRISMA-ScR checklist in \autoref{sec:appprisma}.}. We manually annotate each paper's \textit{Topic Dimension} (Data, Metric, Hypothesis, Reporting, or General) and \textit{Area} (e.g., natural language generation ,machine translation, etc.), alongside a summary and its central methodological position. \autoref{fig:acl-papers} shows paper counts by dimension and publication period, \autoref{fig:area-papers} shows the percentage of papers within each dimension and research area. \autoref{tab:collection-dimensions} compares dimension counts within the ACL and broader ML collections. 

\begin{table}[t]
    \centering
    \small
    \setlength{\tabcolsep}{3pt}
    \renewcommand{\arraystretch}{1.15}
    \begin{tabular*}{\columnwidth}{
        @{\extracolsep{\fill}}lrr@{}
    }
        \toprule
        \textbf{Dimension}
            & \textbf{ACL}
            & \textbf{Broader ML} \\
        & ($n=192$) & ($n=65$) \\
        \midrule
        Data       & 48 (25.0\%) & 29 (44.6\%) \\
        Metrics    & 90 (46.9\%) & 12 (18.5\%) \\
        Hypothesis & 42 (21.9\%) & 27 (41.5\%) \\
        Reporting  & 17 (8.9\%)  & 7 (10.8\%)  \\
        \bottomrule
    \end{tabular*}
    \caption{Taxonomy dimensions represented in the
    ACL and broader ML collections, with entries
    showing paper counts and percentages within each
    collection.}
    \label{tab:collection-dimensions}
\end{table}

\paragraph{Taxonomy Construction}
We developed the taxonomy by manually reviewing each included paper and summarizing its evaluation concerns. Through iterative coding and thematic synthesis \cite{thomas2008methods}, we grouped related concerns and refined their scope through discussion among the authors. This process yielded 22 concerns with concise definitions, organized into four dimensions: \textit{data}, \textit{metrics}, \textit{hypothesis}, and \textit{reporting} (\autoref{fig:unified-taxonomy}). We consolidated these definitions into a codebook to guide consistent interpretation and distinguish related concerns, in line with recommendations for explicit annotation schemes \cite{klie-etal-2024-analyzing}.

\paragraph{Corpus Annotation}
Following research on LLM-based annotation \cite{tan-etal-2024-large}, we used AnnoLLM's \cite{he-etal-2024-annollm} explain-then-annotate procedure with two LLM annotators to assign concern tags using our codebook and labeled demonstrations with explanations. Disagreements were routed to an additional LLM judge following the CLEV framework \cite{badshah-etal-2025-clev}, which adjudicated tags against the source papers. We manually validate the annotations for all 257 papers, checking support for assigned tags and identifying omitted concerns, following process of MEGAnno+ procedure \cite{kim-etal-2024-meganno}. Corrections are documented and ambiguous assignments resolved through discussion among the authors for finalizing tag set for each paper.\footnote{Full annotation and validation procedures are detailed in \autoref{sec:annotation-details}.}

\begin{figure*}[t]
\centering
\resizebox{\textwidth}{!}{%
\begin{tikzpicture}[
  tax node/.style={
    rounded corners=3pt,
    align=center,
    line width=0.45pt,
    inner xsep=4pt,
    inner ysep=2pt,
    minimum height=0.50cm
  },
  tax dimension/.style={
    tax node,
    text width=3.10cm,
    font=\sffamily\bfseries\fontsize{10}{11}\selectfont
  },
  tax category/.style={
    tax node,
    text width=3.00cm,
    font=\sffamily\bfseries\fontsize{8}{9}\selectfont
  },
  tax concern/.style={
    tax node,
    text width=2.70cm,
    font=\sffamily\fontsize{8}{9}\selectfont
  },
  tax arrow/.style={
    line width=0.55pt,
    rounded corners=2pt,
    -{Latex[length=1.5mm,width=1.0mm]}
  }
]

% Root
\node[
  tax node,
  draw=gray!55,
  fill=gray!15,
  text width=3.6cm,
  font=\sffamily\bfseries\fontsize{11}{12}\selectfont
] (taxonomy-root) at (6.15,0) {Evaluation Concerns};

% ============================================================
% DATA
% ============================================================
\begin{scope}[xshift=0.00cm]

\node[tax dimension,
  draw=plasma1!60, fill=plasma1!25, text=plasma1!90!black]
  (dimension-3) at (0,-0.95) {Data};

% Dataset Quality
\node[tax category, draw=plasma1!45, fill=plasma1!10]
  (category-3-2) at (0.10,-1.95) {Dataset\\Quality};
\draw[tax arrow, draw=plasma1!55]
  (dimension-3.south)
  -- (0,-1.40) -- (-1.85,-1.40) |- (category-3-2.west);

\node[tax concern, draw=plasma1!35, fill=plasma1!4]
  (concern-3-2-0) at (0.34,-2.79) {Construct Validity};
\draw[tax arrow, draw=plasma1!45]
  (category-3-2.south west)
  -- ++(0,-0.12) |- (concern-3-2-0.west);

\node[tax concern, draw=plasma1!35, fill=plasma1!4]
  (concern-3-2-1) at (0.34,-3.63) {Gold Standard\\Validity};
\draw[tax arrow, draw=plasma1!45]
  (category-3-2.south west)
  -- ++(0,-0.12) |- (concern-3-2-1.west);

\node[tax concern, draw=plasma1!35, fill=plasma1!4]
  (concern-3-2-2) at (0.34,-4.47) {Artifact Issues};
\draw[tax arrow, draw=plasma1!45]
  (category-3-2.south west)
  -- ++(0,-0.12) |- (concern-3-2-2.west);

\node[tax concern, draw=plasma1!35, fill=plasma1!4]
  (concern-3-2-3) at (0.34,-5.31) {Contamination Issues};
\draw[tax arrow, draw=plasma1!45]
  (category-3-2.south west)
  -- ++(0,-0.12) |- (concern-3-2-3.west);

% Dataset Distribution
\node[tax category, draw=plasma1!45, fill=plasma1!10]
  (category-3-1) at (0.10,-6.37) {Dataset\\Distribution};
\draw[tax arrow, draw=plasma1!55]
  (dimension-3.south)
  -- (0,-1.40) -- (-1.85,-1.40) |- (category-3-1.west);

\node[tax concern, draw=plasma1!35, fill=plasma1!4]
  (concern-3-1-0) at (0.34,-7.21) {Test-Train Splits};
\draw[tax arrow, draw=plasma1!45]
  (category-3-1.south west)
  -- ++(0,-0.12) |- (concern-3-1-0.west);

\node[tax concern, draw=plasma1!35, fill=plasma1!4]
  (concern-3-1-1) at (0.34,-8.05) {Random vs.\\Adversarial};
\draw[tax arrow, draw=plasma1!45]
  (category-3-1.south west)
  -- ++(0,-0.12) |- (concern-3-1-1.west);

% Dataset Assumptions
\node[tax category, draw=plasma1!45, fill=plasma1!10]
  (category-3-0) at (0.10,-9.11) {Dataset\\Assumptions};
\draw[tax arrow, draw=plasma1!55]
  (dimension-3.south)
  -- (0,-1.40) -- (-1.85,-1.40) |- (category-3-0.west);

\node[tax concern, draw=plasma1!35, fill=plasma1!4]
  (concern-3-0-0) at (0.34,-9.95)
  {Empirical vs.\\Rational Stance};
\draw[tax arrow, draw=plasma1!45]
  (category-3-0.south west)
  -- ++(0,-0.12) |- (concern-3-0-0.west);

\end{scope}

\draw[tax arrow, draw=plasma1!55]
  (taxonomy-root.south)
  -- (6.15,-0.48) -| (dimension-3.north);

% ============================================================
% METRICS
% ============================================================
\begin{scope}[xshift=4.10cm]

\node[tax dimension,
  draw=plasma2!60, fill=plasma2!25, text=plasma2!90!black]
  (dimension-2) at (0,-0.95) {Metrics};

% Metric Validity
\node[tax category, draw=plasma2!45, fill=plasma2!10]
  (category-2-2) at (0.10,-1.95) {Metric\\Validity};
\draw[tax arrow, draw=plasma2!55]
  (dimension-2.south)
  -- (0,-1.40) -- (-1.85,-1.40) |- (category-2-2.west);

\node[tax concern, draw=plasma2!35, fill=plasma2!4]
  (concern-2-2-0) at (0.34,-2.79) {Downstream Validity};
\draw[tax arrow, draw=plasma2!45]
  (category-2-2.south west)
  -- ++(0,-0.12) |- (concern-2-2-0.west);

\node[tax concern, draw=plasma2!35, fill=plasma2!4]
  (concern-2-2-1) at (0.34,-3.63) {Linguistic Validity};
\draw[tax arrow, draw=plasma2!45]
  (category-2-2.south west)
  -- ++(0,-0.12) |- (concern-2-2-1.west);

\node[tax concern, draw=plasma2!35, fill=plasma2!4]
  (concern-2-2-2) at (0.34,-4.47) {Human Evaluation};
\draw[tax arrow, draw=plasma2!45]
  (category-2-2.south west)
  -- ++(0,-0.12) |- (concern-2-2-2.west);

% Metric Sensitivity
\node[tax category, draw=plasma2!45, fill=plasma2!10]
  (category-2-1) at (0.10,-5.53) {Metric\\Sensitivity};
\draw[tax arrow, draw=plasma2!55]
  (dimension-2.south)
  -- (0,-1.40) -- (-1.85,-1.40) |- (category-2-1.west);

\node[tax concern, draw=plasma2!35, fill=plasma2!4]
  (concern-2-1-0) at (0.34,-6.37) {Data Sensitivity};
\draw[tax arrow, draw=plasma2!45]
  (category-2-1.south west)
  -- ++(0,-0.12) |- (concern-2-1-0.west);

\node[tax concern, draw=plasma2!35, fill=plasma2!4]
  (concern-2-1-1) at (0.34,-7.21) {Task Sensitivity};
\draw[tax arrow, draw=plasma2!45]
  (category-2-1.south west)
  -- ++(0,-0.12) |- (concern-2-1-1.west);

\node[tax concern, draw=plasma2!35, fill=plasma2!4]
  (concern-2-1-2) at (0.34,-8.05) {Model Sensitivity};
\draw[tax arrow, draw=plasma2!45]
  (category-2-1.south west)
  -- ++(0,-0.12) |- (concern-2-1-2.west);

% Metric Standardization
\node[tax category, draw=plasma2!45, fill=plasma2!10]
  (category-2-0) at (0.10,-9.11) {Metric\\Standardization};
\draw[tax arrow, draw=plasma2!55]
  (dimension-2.south)
  -- (0,-1.40) -- (-1.85,-1.40) |- (category-2-0.west);

\end{scope}

\draw[tax arrow, draw=plasma2!55]
  (taxonomy-root.south)
  -- (6.15,-0.48) -| (dimension-2.north);

% ============================================================
% HYPOTHESIS
% ============================================================
\begin{scope}[xshift=8.20cm]

\node[tax dimension,
  draw=plasma3!60, fill=plasma3!25, text=plasma3!90!black]
  (dimension-1) at (0,-0.95) {Hypothesis};

% Hypothesis Formulation
\node[tax category, draw=plasma3!45, fill=plasma3!10]
  (category-1-2) at (0.10,-1.95) {Hypothesis\\Formulation};
\draw[tax arrow, draw=plasma3!55]
  (dimension-1.south)
  -- (0,-1.40) -- (-1.85,-1.40) |- (category-1-2.west);

\node[tax concern, draw=plasma3!35, fill=plasma3!4]
  (concern-1-2-0) at (0.34,-2.79) {Relevance of\\Hypothesis};
\draw[tax arrow, draw=plasma3!45]
  (category-1-2.south west)
  -- ++(0,-0.12) |- (concern-1-2-0.west);

\node[tax concern, draw=plasma3!35, fill=plasma3!4]
  (concern-1-2-1) at (0.34,-3.63)
  {Task-Specificity\\of Hypothesis};
\draw[tax arrow, draw=plasma3!45]
  (category-1-2.south west)
  -- ++(0,-0.12) |- (concern-1-2-1.west);

% Hypothesis Testing
\node[tax category, draw=plasma3!45, fill=plasma3!10]
  (category-1-1) at (0.10,-4.69) {Hypothesis\\Testing};
\draw[tax arrow, draw=plasma3!55]
  (dimension-1.south)
  -- (0,-1.40) -- (-1.85,-1.40) |- (category-1-1.west);

\node[tax concern, draw=plasma3!35, fill=plasma3!4]
  (concern-1-1-0) at (0.34,-5.53)
  {Reliability of\\Statistical Tests};
\draw[tax arrow, draw=plasma3!45]
  (category-1-1.south west)
  -- ++(0,-0.12) |- (concern-1-1-0.west);

\node[tax concern, draw=plasma3!35, fill=plasma3!4]
  (concern-1-1-1) at (0.34,-6.37)
  {Parameters of\\Statistical Tests};
\draw[tax arrow, draw=plasma3!45]
  (category-1-1.south west)
  -- ++(0,-0.12) |- (concern-1-1-1.west);

% Model Comparison
\node[tax category, draw=plasma3!45, fill=plasma3!10]
  (category-1-0) at (0.10,-7.43) {Model\\Comparison};
\draw[tax arrow, draw=plasma3!55]
  (dimension-1.south)
  -- (0,-1.40) -- (-1.85,-1.40) |- (category-1-0.west);

\node[tax concern, draw=plasma3!35, fill=plasma3!4]
  (concern-1-0-0) at (0.34,-8.27) {Experimental Design};
\draw[tax arrow, draw=plasma3!45]
  (category-1-0.south west)
  -- ++(0,-0.12) |- (concern-1-0-0.west);

\node[tax concern, draw=plasma3!35, fill=plasma3!4]
  (concern-1-0-1) at (0.34,-9.11) {Leaderboard\\Practices};
\draw[tax arrow, draw=plasma3!45]
  (category-1-0.south west)
  -- ++(0,-0.12) |- (concern-1-0-1.west);

\end{scope}

\draw[tax arrow, draw=plasma3!55]
  (taxonomy-root.south)
  -- (6.15,-0.48) -| (dimension-1.north);

% ============================================================
% REPORTING
% ============================================================
\begin{scope}[xshift=12.30cm]

\node[tax dimension,
  draw=plasma4!60, fill=plasma4!25, text=plasma4!90!black]
  (dimension-0) at (0,-0.95) {Reporting};

\node[tax category, draw=plasma4!45, fill=plasma4!10]
  (category-0-1) at (0.10,-1.95) {Transparency\\Concerns};
\draw[tax arrow, draw=plasma4!55]
  (dimension-0.south)
  -- (0,-1.40) -- (-1.85,-1.40) |- (category-0-1.west);

\node[tax category, draw=plasma4!45, fill=plasma4!10]
  (category-0-0) at (0.10,-3.01) {Reproducibility\\Concerns};
\draw[tax arrow, draw=plasma4!55]
  (dimension-0.south)
  -- (0,-1.40) -- (-1.85,-1.40) |- (category-0-0.west);

\end{scope}

\draw[tax arrow, draw=plasma4!55]
  (taxonomy-root.south)
  -- (6.15,-0.48) -| (dimension-0.north);

\end{tikzpicture}%
}

\caption{A taxonomy of evaluation concerns in NLP across
four dimensions: \textcolor{plasma1}{\textbf{Data}},
\textcolor{plasma2}{\textbf{Metrics}},
\textcolor{plasma3}{\textbf{Hypothesis}}, and
\textcolor{plasma4}{\textbf{Reporting}}.
Terminal nodes represent the 22 annotated concerns;
intermediate nodes group related concerns.}
\label{fig:unified-taxonomy}
\end{figure*}

\section{Data Concerns}

%Many critiques of evaluation ultimately trace back to the data used for evaluating systems. We organize data-related evaluation concerns into four high-level categories, reflecting distinct ways in which data can impact model evaluation.

\subsection{Dataset Quality}

These concerns examine the quality of data used to evaluate models.%- how can we ensure the data is good? What factors can create problems in the data? 

\paragraph{Construct Validity}
It concerns whether evaluation tasks and datasets measure the intended capabilities. Reported problems include ambiguous constructs \cite{blodgett-etal-2021-stereotyping, pmlr-v235-zhao24a}, underspecified tasks or insufficient coverage \cite{kryscinski-etal-2019-neural, martinez2023evaluating}, relevance of individual test items \cite{lalor-etal-2016-building, martinez2019item}, broad interpretations of benchmark performance \cite{schlangen-2021-targeting, NEURIPSDATASETSANDBENCHMARKS2021_084b6fbb, tedeschi-etal-2023-whats}, increasing reliance on a narrow set of benchmarks \cite{paullada2021data, gururangan-etal-2022-whose, koch2021reduced}, or insufficient documentation of datasets \cite{10.1145/3442188.3445918, bhardwaj2024machine}.

\noindent$\rightarrow$ {\em Proposed responses include unit-test inspired testing \cite{ribeiro-etal-2020-beyond, saxon2024benchmarks}, psychometric modeling \cite{lalor-etal-2016-building, martinez2019item, sedoc-ungar-2020-item}, and systematic dataset curation and benchmark assessment \cite{bhardwaj2024machine, 10.5555/3737916.3740660, reuel2024betterbench}.}

\paragraph{Gold Standard Validity}
Works discuss whether reference annotations provide defensible ground truth. Specific concerns include label errors and inconsistencies \cite{manning2011part, northcutt2021pervasive}, uneven annotator expertise \cite{whitehill2009whose, pmlr-v9-yan10a}, exploitable annotator biases \cite{geva-etal-2019-modeling}, genuine ambiguity of annotations \cite{basile-etal-2021-need, fleisig-etal-2024-perspectivist}, and expert-annotation costs \cite{snow-etal-2008-cheap, hovy-etal-2014-experiments, louis-nenkova-2013-automatically}.

\noindent$\rightarrow$ {\em Research has suggested span label auditing \cite{manning2011part, northcutt2021pervasive}, repeated labeling with expertise-aware aggregation \cite{10.1145/1401890.1401965, whitehill2009whose, welinder2010multidimensional}, and annotator filtering \cite{JMLR:v13:raykar12a} to mitigate these issues. Other work incorporates disagreement into modeling and evaluation \cite{plank-etal-2014-learning, davani-etal-2022-dealing, baan-etal-2022-stop}.}

\paragraph{Artifact Issues}
Artifact issues arise when specific cues enable model success without the intended reasoning. Examples include hypothesis-only label cues \cite{tsuchiya-2018-performance, gururangan-etal-2018-annotation}, shallow lexical and syntactic heuristics \cite{mccoy-etal-2019-right, du-etal-2021-towards}, and document-layout biases \cite{kryscinski-etal-2019-neural}. Such shortcuts conceal failures under lexical or distribution shifts \cite{glockner-etal-2018-breaking, clark-etal-2019-dont}.

\noindent$\rightarrow$ {\em  Controlled challenge sets \cite{glockner-etal-2018-breaking, mccoy-etal-2019-right}, premise-sensitive training objectives \cite{belinkov-etal-2019-dont}, ensembles accounting for known biases \cite{clark-etal-2019-dont}, and regularization against shortcut learning \cite{du-etal-2021-towards} have been considered as mitigations.}

\paragraph{Contamination Issues}
Contamination occurs when evaluation data enters model training, potentially inflating scores and compromising comparisons. Undisclosed training corpora make exposure difficult to establish and memorization difficult to distinguish from generalization.

\noindent$\rightarrow$ {\em Addressing contamination involves detecting exposure for each model--benchmark pair \cite{sainz-etal-2023-nlp} or shuffled benchmark orderings to detect contamination \cite{oren2024proving}.}

\subsection{Dataset Distribution} 

These concerns are about how the distributions of evaluation data should relate to training data or real-world inputs.

\paragraph{Test-Train Splits}
Test-train splits determine which forms of generalization an evaluation measures. Concerns include dependencies across partitions \cite{karimi-etal-2015-squibs, finegan-dollak-etal-2018-improving}, mismatches with intended deployment conditions \cite{nie-etal-2022-impact, NEURIPS2023_b6b5f50a}, rankings sensitive to split or test-set composition \cite{szymanski-gorman-2020-best, kovatchev-lease-2024-benchmark}, and overfitting through repeated development- and test-set use \cite{van-der-goot-2021-need}.

\noindent$\rightarrow$ {\em Suggested safeguards include query-based or chronological partitions \cite{finegan-dollak-etal-2018-improving, nie-etal-2022-impact}, explicit control of evaluation distributions \cite{kovatchev-lease-2024-benchmark, NEURIPS2023_b6b5f50a}, cross-validated model comparisons \cite{szymanski-gorman-2020-best}, and separate tuning and development sets \cite{van-der-goot-2021-need}.}

\paragraph{Random vs.\ Adversarial}
This concern examines whether evaluation samples typical inputs or deliberately probes failure cases. Random splits can yield optimistic estimates and poorly represent deployment distributions \cite{sogaard-etal-2021-need, siska-etal-2024-examining} while adversarial evaluations reveal weaknesses in linguistic reasoning \cite{isabelle-etal-2017-challenge, naik-etal-2018-stress}, robustness to input perturbations \cite{belinkov2018synthetic, moradi-samwald-2021-evaluating, wallace-etal-2019-universal}, and generalization \cite{agrawal-etal-2023-reassessing, wang2023on}.

\noindent$\rightarrow$ {\em Evaluation splits should be chosen based on purpose of evaluation and there is no one right answer for making such decisions \cite{gorman-bedrick-2019-need, rajpurkar-etal-2018-know, gardner-etal-2020-evaluating, kiela-etal-2021-dynabench}.}

\subsection{Dataset Assumptions}

Rather than focusing on specific dataset pathologies, some works have examined the assumptions underlying benchmark-driven evaluation. 

\paragraph{Empirical vs.\ Rational Stance}
This concern contrasts corpus-based evidence about observed performance with theory-driven accounts of linguistic competence \cite{church2011pendulum, church2019survey}. Critiques warn that prioritizing benchmark gains can marginalize linguistic theory and produce increasingly strong results with limited explanation of what models learn or why they succeed \cite{church2017emerging, church2019survey}.

\noindent$\rightarrow$ {\em These accounts advocate drawing on both empirical and rationalist perspectives \cite{church2011pendulum, church2019survey}, linking quantitative evaluation to linguistic analysis and explanatory insight \cite{church2017emerging, church2019survey}.}

\section{Metric Concerns}

\subsection{Metric Validity}

Widely adopted metrics can abstract away from the underlying phenomena of interest leading to evaluations that are substantively misaligned with real-world or theoretical goals.

\paragraph{Downstream Validity}
This asks whether metric gains translate into better end-task performance or practical utility. Reported gaps include weak correspondence with human quality judgments \cite{callison-burch-etal-2006-evaluating, reiter-belz-2009-investigation, novikova-etal-2017-need}, poor prediction of downstream performance \cite{headden-iii-etal-2008-evaluating}, and misalignment with user needs \cite{tahri-etal-2023-transitioning, 10.5555/3042573.3042809}. Similar aggregate scores can also conceal substantial differences in system behavior \cite{dutta2024accuracy}.

\noindent$\rightarrow$ {\em Validation efforts use downstream-task testing \cite{headden-iii-etal-2008-evaluating} or user-satisfaction modeling \cite{walker-etal-2001-quantitative}, alongside application-oriented measures \cite{mccowan2004use} and estimators combining automatic metrics with human judgments \cite{chaganty-etal-2018-price}.}

\paragraph{Linguistic Validity}
It asks whether metrics capture relevant linguistic properties. Surface overlap can miss syntactic or semantic distinctions \cite{liu-gildea-2005-syntactic, pado-etal-2009-robust, guzman-etal-2014-using}, summary-content scores can overlook linguistic quality and factual consistency \cite{conroy-dang-2008-mind, kryscinski-etal-2019-neural}. Metrics may also penalize valid alternative analyses or corrections \cite{plank-etal-2014-linguistically, napoles-etal-2016-theres}, and reward questionable changes through their aggregation behavior \cite{chiang-etal-2008-decomposability}.

\noindent$\rightarrow$ {\em Richer scoring methods incorporate syntax \cite{liu-gildea-2005-syntactic}, semantic information \cite{fum-etal-1985-rule, pado-etal-2009-robust}, discourse relations \cite{guzman-etal-2014-using}, reference-free grammaticality \cite{napoles-etal-2016-theres}, and decomposable scoring \cite{chiang-etal-2008-decomposability}, although errors in automated linguistic analysis can limit their reliability \cite{gimenez-marquez-2009-robustness}.}

\paragraph{Human Evaluation}
Human evaluation concerns whether human judgments provide valid, reliable assessments. Challenges include inconsistent criteria and reporting \cite{howcroft-etal-2020-twenty}, differences in evaluator expertise and cognitive biases \cite{clark-etal-2021-thats, karpinska-etal-2021-perils, elangovan-etal-2024-considers}, sensitivity to rating scales and agreement measures \cite{belz-kow-2011-discrete, amidei-etal-2019-agreement}, annotation workload \cite{krishna-etal-2023-longeval}, and unstable rankings or flawed experimental execution \cite{van-miltenburg-etal-2023-reproducible, riley-etal-2024-finding, thomson2024common}.

\noindent$\rightarrow$ {\em Efforts to strengthen human evaluation emphasize standardized criteria and reporting \cite{howcroft-etal-2020-twenty}, reference-based rater calibration \cite{karpinska-etal-2021-perils}, finer-grained judgments \cite{krishna-etal-2023-longeval}, complementary agreement and correlation measures \cite{amidei-etal-2019-agreement}, ranking-stability analysis \cite{riley-etal-2024-finding}, and preregistration and piloting \cite{thomson2024common}.}

\subsection{Metric Sensitivity}

Another set of concerns focuses on how metrics are sensitive to properties of the evaluation setting such as tasks, datasets, or model classes.

\paragraph{Data Sensitivity}
Metric scores and their agreement with human judgments can depend on reference choice \cite{donaway-etal-2000-comparison}, dataset size and outliers \cite{reichart-rappoport-2009-nvi, mathur-etal-2020-tangled}, domain and language \cite{bhandari-etal-2020-evaluating, forde-etal-2024-evaluating}, and lexical composition \cite{belem2024are}.

\noindent$\rightarrow$ {\em Responses include size-normalized measures \cite{reichart-rappoport-2009-nvi}, validation against human judgments on target datasets \cite{bhandari-etal-2020-evaluating, forde-etal-2024-evaluating}, and explicit measures of cross-domain robustness \cite{yu-etal-2022-measuring}.}

\paragraph{Task Sensitivity}
Metrics transferred across tasks can correlate poorly with human judgments \cite{elliott-keller-2014-comparing, liu-etal-2016-evaluate}, overlook task goals and complexity \cite{thomas-1999-designing, babych-etal-2004-extending}, or miss contextual relevance \cite{muller-etal-2014-predicting, thomson-reiter-2021-generation}.

\noindent$\rightarrow$ {\em Task-specific metric designs can account for goal priorities \cite{thomas-1999-designing}, normalize translation scores for complexity \cite{babych-etal-2004-extending}, and assess dialogue coherence through entailment \cite{dziri-etal-2019-evaluating}, with metric selection informed by human judgments \cite{elliott-keller-2014-comparing}.}

\paragraph{Model Sensitivity}
Metrics can favor frequency-based generators \cite{belz-reiter-2006-comparing}, outputs resembling their underlying models \cite{deutsch-etal-2022-limitations, liu-etal-2024-llms-narcissistic}, or machine over human text \cite{kasai-etal-2022-bidimensional}. LLM judges also exhibit position and verbosity biases \cite{10.5555/3666122.3668142}, while stereotype metrics can overlook variability across generations \cite{10.5555/3737916.3741411}.

\noindent$\rightarrow$ {\em Mitigation strategies explore human-validated metric ensembles \cite{kasai-etal-2022-bidimensional}, order-swapped and reference-guided judgments \cite{10.5555/3666122.3668142}, and distributional estimates capturing generation variability \cite{10.5555/3737916.3741411}.}

\subsection{Metric Standardization}
Metric standardization concerns justified metric selection and use. Problems include habitual adoption despite known weaknesses \cite{caglayan-etal-2020-curious}, inconsistent definitions and protocols \cite{han-etal-2023-fair, belz-etal-2020-disentangling}, missing implementation details \cite{schmidtova-etal-2024-automatic-metrics}, and insufficient validation of metric variants \cite{graham-2015-evaluating}. Reliance on single scores can also obscure distinct quality dimensions \cite{lee-etal-2024-comparative}.

\noindent$\rightarrow$ {\em To improve comparability, studies propose shared evaluation tools \cite{fabbri-etal-2021-summeval}, explicit protocol classifications \cite{belz-etal-2020-disentangling}, significance tests and uncertainty estimates for metric validation \cite{graham-baldwin-2014-testing, xiao-etal-2023-evaluating-evaluation}, and combinations of complementary metrics \cite{gimenez-marquez-2008-heterogeneous}.}

\section{Hypothesis Concerns}

\subsection{Hypothesis Formulation}
These concerns address which claims evaluations should test and whether the chosen evaluation setups support those claims.

\paragraph{Relevance of Hypothesis}
Some concerns here include unclear evaluation scope and stakeholder goals \cite{mann-1981-selective, thompson-1981-evaluation}, underspecified concepts \cite{blodgett-etal-2020-language, jacovi-goldberg-2020-towards, fierro-etal-2024-defining}, and prioritizing predictive performance over practical benefits \cite{10.5555/3042573.3042809, 10.1145/3531146.3533233}.

\noindent$\rightarrow$ {\em Suggested approaches clarify task definitions and normative assumptions \cite{subramonian-etal-2023-takes, blodgett-etal-2020-language}, distinguish specification compliance from user requirements \cite{barr-klavans-2001-verification}, and seek converging evidence across multiple capability tests \cite{frank2023baby}.}

\paragraph{Task-Specificity of Hypothesis}
Evaluation tasks can support narrower claims than those made about models: embedding quality varies across tasks \cite{sahlgren-2006-towards, schnabel-etal-2015-evaluation}, bilingual lexicon induction can misrepresent downstream usefulness \cite{glavas-etal-2019-properly}, and benchmark success need not imply human-like generalization \cite{linzen-2020-accelerate}.

\noindent$\rightarrow$ {\em Closer alignment uses evaluations situated in actual user tasks \cite{whittaker-stenton-1989-user}, comparisons across downstream applications \cite{glavas-etal-2019-properly}, and direct tests of linguistic properties beyond perplexity \cite{meister-cotterell-2021-language}.}

\subsection{Hypothesis Testing}
These concerns address whether statistical evidence supports performance claims and how test choices affect conclusions.

\paragraph{Reliability of Statistical Tests}
Significance claims can be distorted by violated independence assumptions \cite{dietterich1998approximate, yeh-2000-accurate}, optimizer instability \cite{clark-etal-2011-better}, and uncorrected multiple comparisons \cite{dror-etal-2017-replicability}. 

\noindent$\rightarrow$ {\em Proposed safeguards include design-appropriate paired or resampling tests \cite{koehn-2004-statistical, dror-etal-2018-hitchhikers}, repeated optimization runs \cite{clark-etal-2011-better}, and procedures accounting for comparisons across multiple datasets \cite{10.5555/1248547.1248548, dror-etal-2017-replicability}.}

\paragraph{Parameters of Statistical Tests}
Conventional significance thresholds can be too permissive under flexible evaluation choices \cite{sogaard-etal-2014-whats}, while small samples reduce power and exaggerate detected effects \cite{card-etal-2020-little}. Significance and effect-size estimates also depend on sample composition, system similarity, and domain shifts \cite{berg-kirkpatrick-etal-2012-empirical, sogaard-2013-estimating}.

\noindent$\rightarrow$ {\em Work on parameter choice advocates power-based sample sizing \cite{card-etal-2020-little}, more stringent significance thresholds \cite{sogaard-etal-2014-whats}, and meta-analysis of effect sizes across diverse datasets \cite{sogaard-2013-estimating}.}

\subsection{Model Comparison}
These concerns address whether experimental comparisons and rankings justify claims that one system is better.

\paragraph{Experimental Design}
Comparisons can be confounded by test-set reuse \cite{salzberg1997comparing, liao2021we}, entangled modeling choices \cite{klein-manning-2002-conditional}, and differences in training-data size or preprocessing \cite{ng2001discriminative, 10.5555/3737916.3740955}. Aggregate scores can also hide complementary system strengths \cite{derczynski-2016-complementarity}.

\noindent$\rightarrow$ {\em Stronger designs isolate modeling factors \cite{klein-manning-2002-conditional}, separate tuning from testing \cite{salzberg1997comparing}, and use of appropriate cross-validation procedures \cite{arlot2010survey, xue-etal-2023-need} to improve comparison reliability.}

\paragraph{Leaderboard Practices}
Leaderboards can encourage adaptive overfitting \cite{blum2015ladder}, obscure differences in item informativeness \cite{rodriguez-etal-2021-evaluation}, and support misleading claims of superhuman performance \cite{tedeschi-etal-2023-whats}. Aggregating incompatible criteria \cite{10.5555/3648699.3648930} and prioritizing scores over explanation \cite{church2017emerging, church2019survey} further weaken their interpretation.

\noindent$\rightarrow$ {\em Alternatives include the Ladder algorithm for adaptive overfitting \cite{blum2015ladder}, item response theory for modeling item difficulty and discrimination \cite{rodriguez-etal-2021-evaluation}, and statistically tested stochastic-dominance comparisons across multiple criteria \cite{10.5555/3648699.3648930}.}

\section{Reporting Concerns}
These concerns address whether evaluation reports enable meaningful interpretation and reproduction.

\subsection{Transparency Concerns}
Score-only reporting can conceal development budgets \cite{dodge-etal-2019-show}, scoring choices \cite{palen-michel-etal-2021-seqscore}, and failures on individual examples \cite{burnell2023rethink}. Missing information about practical costs \cite{ethayarajh-jurafsky-2020-utility}, measurement assumptions \cite{goldfarb-tarrant-etal-2023-prompt}, and stakeholders or contexts of use \cite{liu-etal-2023-responsible} further limits interpretation.

\noindent$\rightarrow$ {\em More informative reporting provides performance--budget curves \cite{dodge-etal-2019-show}, instance-level results \cite{burnell2023rethink}, explicit scoring documentation \cite{palen-michel-etal-2021-seqscore}, and benchmark checklists covering assumptions and limitations \cite{reuel2024betterbench}.}

\subsection{Reproducibility Concerns}
Incomplete documentation and inaccessible code or data hinder reproduction \cite{storks-etal-2023-nlp}, particularly for human evaluations \cite{belz-etal-2023-missing, belz-etal-2023-non}. Further problems include inconsistent definitions \cite{belz2022metrological}, experimental errors \cite{thomson2024common}, and software bugs that persist even when results reproduce \cite{papi-etal-2024-good}.

\noindent$\rightarrow$ {\em Practical responses combine accessible research materials \cite{storks-etal-2023-nlp}, workflow management \cite{digan2021can}, software testing and code-quality checklists \cite{papi-etal-2024-good}, and quantified assessments supported by coordinated reproduction studies \cite{belz-etal-2022-quantified, belz-thomson-2024-2024}.}

\begin{figure}[t]
    \centering
    \includegraphics[width=\columnwidth]
        {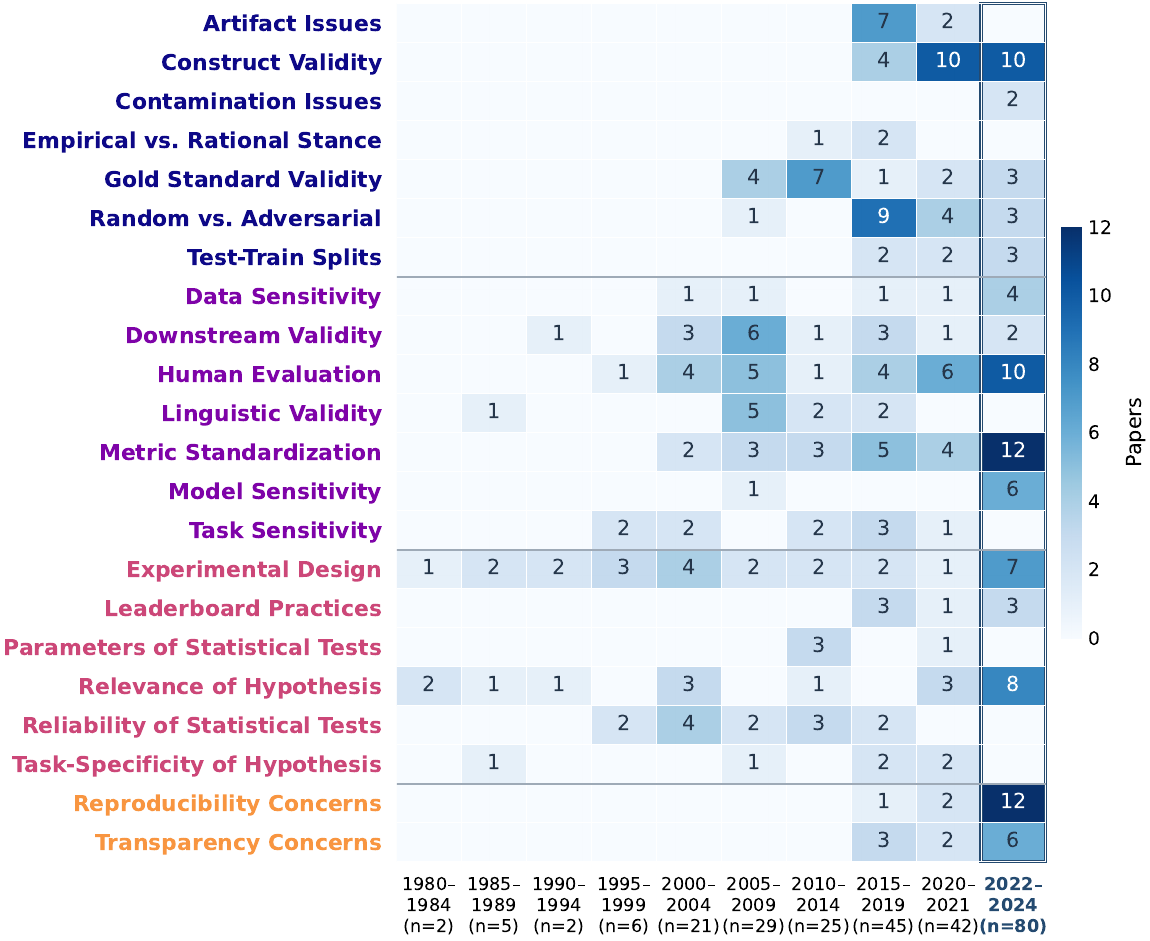}
    \caption{Temporal coverage of the 22 evaluation concerns in the 257-paper corpus. Cells show tagged-paper counts; blank cells indicate zero papers. Label colours follow the taxonomy, and period totals appear below the horizontal axis.}
    \label{fig:temporal-tags}
\end{figure}

\begin{figure}[t]
    \centering
    \includegraphics[width=\columnwidth]
        {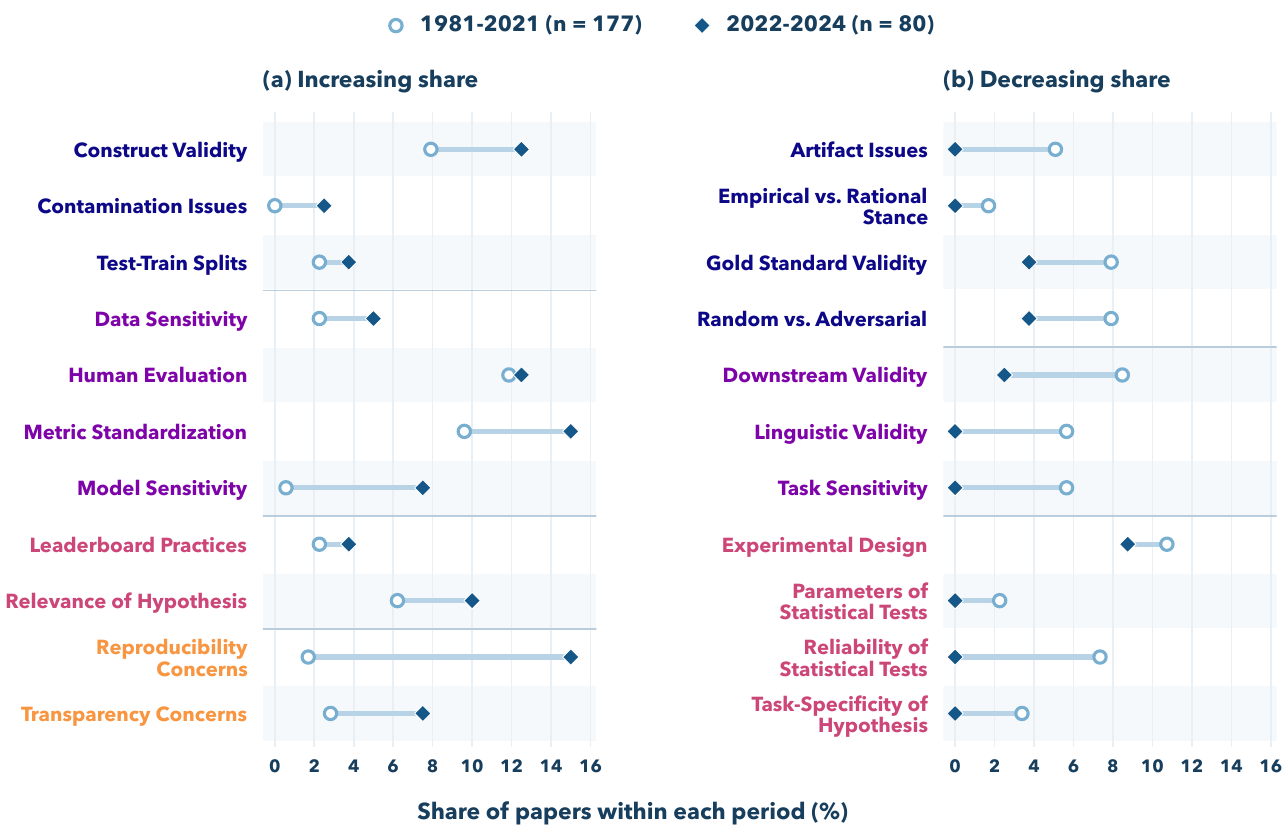}
    \caption{Changes in the share of papers addressing each concern. Light-blue circles represent 1981--2021 ($n=177$); dark-blue diamonds represent 2022--2024 ($n=80$). Panels show the 11 concerns with increasing share (a) and the 11 with decreasing share (b), using the same scale.}
    \label{fig:concern-prevalence-change}
\end{figure}

\section{Current Status } % Where We Stand
\label{sec:concernsstatus}

We examine the current status of the 22 concerns: first, we use our corpus to trace when each concern was last discussed and how the volume and distribution of attention have changed over time. Second, we examine institutionalization efforts, including the ACL Responsible NLP Research and NeurIPS checklists, identifying which concerns receive explicit methodological guidance or requests for disclosure.

\subsection{Recency Analysis}
\label{subsec:recency-analysis}

We use the final annotations to examine when each concern was last represented in the corpus and how its relative share of papers has changed.

\paragraph{\textbf{Recency of discussion.}}  \autoref{fig:temporal-tags} traces the 22 concerns across publication periods. We consider 2022--2024, the final three years covered by the survey as the recent window. \textit{Reproducibility Concerns}, \textit{Model Sensitivity}, and \textit{Metric Standardization} account for larger shares of recent papers. Seven concerns have no tagged papers in this window: \textit{Artifact Issues}, \textit{Empirical vs.\ Rational Stance}, \textit{Linguistic Validity}, \textit{Task Sensitivity}, \textit{Task-Specificity of Hypothesis}, \textit{Reliability of Statistical Tests}, and \textit{Parameters of Statistical Tests}.

\paragraph {\textbf{Changes in relative attention.}} \autoref{fig:concern-prevalence-change} compares the percentage of papers assigned to each concern in 1981--2021 ($n=177$) and 2022--2024 ($n=80$). The largest increases are in \textit{Reproducibility Concerns} (1.7\% to 15.0\%) and \textit{Model Sensitivity} (0.6\% to 7.5\%), suggesting greater relative attention
to replicating results and understanding model-dependent metric behaviour. \textit{Human Evaluation} remains prominent but comparatively stable (11.9\% to 12.5\%) while \textit{Downstream Validity} declines from 8.5\% to 2.5\% .

To complement this picture of research attention, we next examine which concerns receive explicit recognition in institutional checklists.

\begin{table}[t]
    \centering
    \footnotesize
    \setlength{\tabcolsep}{2pt}
    \renewcommand{\arraystretch}{1.16}

    \begin{tabular*}{\columnwidth}{
        @{\extracolsep{\fill}}
        >{\raggedright\arraybackslash}p{0.52\columnwidth}
        cc@{}
    }
        \toprule
        \textbf{Concern} & \textbf{ACL} & \textbf{NeurIPS} \\
        \midrule

        \rowcolor{plasma1!12}[0pt][0pt]
        \multicolumn{3}{@{}p{\columnwidth}@{}}
            {\strut\textbf{Data}} \\
        Construct Validity & No & No \\
        Gold Standard Validity & No & No \\
        Artifact Issues & No & No \\
        Contamination Issues
            & Yes \textbf{(C2)} & No \\
        Test-Train Splits
            & Yes \textbf{(B6)} & Yes \textbf{(6)} \\
        Random vs.\ Adversarial & No & No \\
        Empirical vs.\ Rational Stance & No & No \\

        \midrule
        \rowcolor{plasma2!12}[0pt][0pt]
        \multicolumn{3}{@{}p{\columnwidth}@{}}
            {\strut\textbf{Metrics}} \\
        Downstream Validity & No & No \\
        Linguistic Validity & No & No \\
        Human Evaluation
            & Yes \textbf{(D1--D2)} & Yes \textbf{(14)} \\
        Data Sensitivity & No & No \\
        Task Sensitivity & No & No \\
        Model Sensitivity & No & No \\
        Metric Standardization
            & Yes \textbf{(C4)} & No \\

        \midrule
        \rowcolor{plasma3!12}[0pt][0pt]
        \multicolumn{3}{@{}p{\columnwidth}@{}}
            {\strut\textbf{Hypothesis}} \\
        Relevance of Hypothesis & No & No \\
        Task-Specificity of Hypothesis
            & Yes \textbf{(A1)} & Yes \textbf{(1--2)} \\
        Reliability of Statistical Tests
            & Yes \textbf{(C3)} & Yes \textbf{(7)} \\
        Parameters of Statistical Tests & No & No \\
        Experimental Design & No & No \\
        Leaderboard Practices & No & No \\

        \midrule
        \rowcolor{plasma4!12}[0pt][0pt]
        \multicolumn{3}{@{}p{\columnwidth}@{}}
            {\strut\textbf{Reporting}} \\
        Transparency Concerns
            & Yes \textbf{(A1, C)} & Yes \textbf{(1--8)} \\
        Reproducibility Concerns
            & Yes \textbf{(C)} & Yes \textbf{(4--6)} \\

        \midrule
        \textbf{Concerns with coverage}
            & \textbf{8/22} & \textbf{6/22} \\
        \bottomrule
    \end{tabular*}

    \caption{Coverage of the 22 evaluation concerns in the ACL Responsible NLP Research and NeurIPS Paper checklists, where Yes indicates an explicit methodological or disclosure
    request addressing a component of the concern, No indicates no targeted item under our coding rule, and bold parenthetical references identify the relevant checklist items or sections.}
    \label{tab:checklist-coverage}
\end{table}

\begin{comment}

\begin{figure}[t]
    \centering
    \includegraphics[width=\columnwidth]
    {images/checklist_coverage_by_dimension.pdf}
    \caption{Explicit checklist coverage across the four taxonomy dimensions, with blue indicating concerns addressed by a targeted checklist item, and fractions showing the number covered out of the total in each dimension. Coverage may address only one
    component of a concern and does not establish its resolution or adoption in practice.}
    \label{fig:checklist-coverage-dimensions}
\end{figure}
\end{comment}

\subsection{Standardization Efforts}
\label{subsec:standardization-efforts}

Research checklists provide an institutional mechanism for translating methodological concerns into questions that authors address at submission. We examine the \textit{ACL Responsible NLP Research Checklist}\footnote{Official checklist and guidance: \url{https://aclrollingreview.org/responsibleNLPresearch/}.} and the \textit{NeurIPS Paper Checklist}\footnote{Official checklist and guidance: \url{https://neurips.cc/public/guides/PaperChecklist}.} for seeing whether they address or ask for clarifications for any of the concerns in our taxonomy.

\paragraph{Checklist analysis}
We map the 22 concerns to explicit checklist items, recording coverage when an item requests methodological action or disclosure addressing a component of a concern. Under this coding, ACL covers 8/22 concerns and NeurIPS 6/22; \autoref{tab:checklist-coverage} provides the supporting item identifiers. Coverage indicates institutional recognition, without establishing adoption or resolution.

\paragraph{Section-wise analysis}
ACL's limitations section (A1) and NeurIPS's claims and limitations items (1--2) ask authors to justify the scope of their conclusions, supporting task-specificity and transparency. Data and experimental-setting items (ACL B6; NeurIPS 6) request train--test split details. Computational-experiment items (ACL C; NeurIPS 4--8) emphasize reproducibility, statistical uncertainty, and resource reporting; ACL additionally discourages tuning on held-out data (C2) and requests metric implementations and settings (C4). Human-participant items (ACL D1--D2; NeurIPS 14) request instructions and compensation details, addressing aspects of human evaluation. These requests emphasize disclosure and procedural safeguards, while 14 concerns lack targeted items in either checklist under our coding. Our complementary checklist (Appendix E, \autoref{fig:eval-checklist}) provides prompts for examining these less-addressed evaluation choices.

\section{Conclusion}

In this work, we bring together the debates and discussions surrounding evaluation methodology to give researchers historical context for assessing current practices and identifying assumptions that require renewed scrutiny. We plan to develop the reviewed works into a living repository of papers that the community can contribute to and use as a shared reference for evaluation concerns.

Moreover, our analysis also reveals uneven attention to these concerns: some remain prominent in research, while others have become less visible in the literature we reviewed. Institutional checklists so far also do not provide exhaustive coverage for all concerns. Together, these patterns highlight opportunities for the research community to revisit unresolved questions and develop clearer guidance for evaluation practice.

We hope this work serves as both a reminder of what the community has done and a prompt to revisit what has been left unresolved.  This  may be one of the most important missions for NLP research going forward, and a shared responsibility as well as a technical challenge for future work.

\section*{Limitations}

We note three limitations of this work. \textit{First}, as with any scoping review, our coverage is shaped by the review's scope and search strategy. We focus on representative rather than exhaustive coverage of the literature, so relevant work may be missing. We plan to develop a living repository that the community can contribute to and consult as a shared reference, allowing coverage to expand over time.

\textit{Second}, the taxonomy reflects subjective judgments made through iterative coding and thematic synthesis. Its purpose is to map the space of evaluation concerns rather than establish absolute divisions in the literature. Some concerns overlap, and alternative classifications may offer complementary perspectives.

\textit{Third}, our analysis focuses on two institutional checklists and may overlook relevant guidance provided elsewhere. Our proposed checklist is intended as a companion for reflection on evaluation choices, rather than a definitive standard or a set of requirements for universal adoption.

\section*{Ethical Considerations}

This study analyzes publicly available research papers. We did not identify offensive content in the reviewed publications or in the annotations produced for this study. All manual screening, taxonomy development, and annotation validation were undertaken by the authors as part of the research. We did not recruit external annotators, crowdworkers, or human subjects; participant recruitment, compensation, and participant-facing consent procedures
are therefore not applicable.

We used AI-based LLMs to annotate and adjudicate concern labels under an author-developed taxonomy, as detailed in \autoref{sec:annotation-details}. Automated annotations may misrepresent a paper's contributions or overlook relevant concerns. To address these risks, the authors manually checked all final annotations against the source papers using the concern codebook and validation procedure. The authors retain responsibility for the interpretations and findings presented.

\begin{comment}
\section*{Acknowledgments}

We will include all relevant acknowledgements in the final published version of the paper. 

\end{comment}


\bibliography{custom}

\appendix

\section{Related Work}
\label{sec:apprelatedwork}

Broad treatments of NLP evaluation provide important foundations for this survey \cite{1457848, jones1995evaluating, king1996evaluating, resnik2010evaluation}. Historical perspectives examine the relationship between empirical performance and linguistic understanding \cite{church2011pendulum}, question the emphasis on improved scores over explanation \cite{church2017emerging}, and trace the rise of shared tasks and leaderboards \cite{church2019survey}. These accounts establish a long history of reflection on what evaluation results mean for progress in NLP. 

More focused reviews have examined particular evaluation concerns in depth. \citet{howcroft-etal-2020-twenty} document inconsistent criteria and reporting in human evaluation of NLG systems. \citet{10.1613/jair.1.13715} synthesize problems with NLG datasets and automatic and human evaluation, organize proposed responses, and examine their uptake in published work. \citet{belz-etal-2021-systematic} review reproducibility research in NLP, including disagreements over its definition and measurement. These studies provide detailed analyses of concerns that recur across our taxonomy. Related syntheses have also examined methodological failures across ML: \citet{liao2021we} organize critiques from multiple subfields into a taxonomy of internal and external validity failures, \citet{10.1145/3317287.3328534} examine weaknesses in scientific argumentation and communication, while \citet{10.1145/3531146.3533233} analyze gaps between narrow predictive evaluations and the requirements of application contexts. \citet{eriksson2025ai} bring together technical and sociotechnical critiques of AI benchmarks, including construct validity and the incentives shaping benchmark use. Recent LLM surveys map evaluation tasks and methods \cite{chang2024survey} and critically review decisions that affect evaluation reliability and reproducibility \cite{laskar2024systematic}. Complementary efforts turn methodological concerns into practical guidance: BetterBench assesses benchmark development and use against best practices \cite{reuel2024betterbench}, while \citet{burnell2023rethink} advocate reporting instance-level results to support more informative analyses of AI performance.

However, no work reviewed above combines a historical synthesis of NLP evaluation concerns with an analysis of their changing prominence and coverage in institutional research checklists. The novelty of our work lies in bringing these perspectives together: we organize four decades of methodological discussion into 22 concerns across \textit{data}, \textit{metrics}, \textit{hypothesis}, and \textit{reporting}, connecting the problems raised to proposed responses. Using our annotated corpus, we examine how these concerns vary across research areas and how attention to them has changed over time. We also map their coverage in the ACL and NeurIPS checklists, identifying concerns that receive limited recent discussion or institutional guidance. We translate this synthesis into a reflective checklist for researchers (\autoref{fig:eval-checklist}) and plan to develop a community-contributed repository to support continued engagement with this methodological history.

\section{Literature Review Methodology}
\label{sec:appsurveymethod}

\paragraph{Scope}
We conducted a scoping review \cite{pham2014scoping, peters2021scoping} to synthesize methodological concerns surrounding evaluation in NLP, informed by relevant work from the broader ML literature. Given the breadth of this literature, we sought representative coverage of recurring concerns and proposed responses rather than exhaustive coverage. We considered publications up to 2024. The PRISMA-ScR reporting checklist \cite{tricco2018prisma} is provided in \autoref{sec:appprisma}. The included publications are written in English. This describes the language of the reviewed documents, not necessarily the languages studied within those publications.

\paragraph{Search strategy}
We combined keyword-based retrieval with human screening against explicit inclusion criteria, as also used in structured reviews of NLP research \cite{belz-etal-2021-systematic}. We searched two bibliographic sources: the \textbf{ACL Anthology}\footnote{\url{https://aclanthology.org}} and \textbf{Semantic Scholar}\footnote{\url{https://www.semanticscholar.org}}. Semantic Scholar extended retrieval to relevant work in broader ML venues, including ICML, AAAI, and ACM and IEEE publications.

For the ACL Anthology, we initially searched for grammatical variants of \textit{evaluation}. We conducted similar searches in Semantic Scholar, filtering by \textit{Fields of Study} (Computer Science and Linguistics), \textit{Date Range} (1980--2024), \textit{PDF Availability} (Yes), and relevant publication venues. We iteratively expanded the search vocabulary to capture related methodological discussions, including variants of \textit{measurement}, \textit{testing}, and \textit{benchmark}. These searches yielded approximately 5000 candidate records, which were screened through human inspection of titles or abstracts.

\paragraph{Screening Strategy}
A paper was included only if it satisfied both of the following criteria:

\begin{enumerate}[label=\textbf{IC\arabic*}, leftmargin=2.2em]
    \item \textbf{Evaluation relevance:}
    The paper explicitly examined evaluation as a methodological concern
    and addressed at least one of the following: limitations or biases
    in evaluation practices; assumptions underlying evaluation design,
    including task, data, or metric choices; misalignment between
    evaluation results and model behavior; or normative or methodological
    arguments about evaluation.

    \item \textbf{Substantive focus:}
    Evaluation concerns formed a central component of the paper's
    contribution rather than a peripheral motivation or discussion
    point. We excluded papers whose primary contribution was a new
    artifact without sustained critical analysis of evaluation practices,
    as well as papers discussing evaluation only incidentally.
\end{enumerate}

Borderline cases were handled conservatively: papers were retained only when evaluation concerns were clearly articulated and substantively developed. This screening stage yielded 194 papers across the ACL and broader ML collections.

\paragraph{Citation-based expansion}
We supplemented the initial search with backward citation screening, examining references cited by the retained papers. Supplementing keyword retrieval through citation links has also been used in critical NLP surveys to identify relevant work beyond the initial search vocabulary \cite{blodgett-etal-2020-language}. We assessed these additional candidates against the same inclusion criteria. This step aimed to improve coverage of influential contributions missed by keyword searches, particularly earlier work and papers using different terminology for evaluation concerns.

The final corpus contained 257 papers: 192 from the ACL Anthology and 65 from the broader ML collection, spanning 1981--2024.

\paragraph{Manual Annotation}
We recorded structured information about each included paper to support comparison and synthesis, an approach also used in surveys of evaluation practice \cite{howcroft-etal-2020-twenty}. Alongside bibliographic metadata, we manually recorded each paper's \textit{Topic} (Data, Metric, Hypothesis, Reporting, or General), \textit{Area} (e.g., machine translation or natural language generation), a summary of its evaluation concerns, and its central methodological position. The following process for taxonomy construction, automated annotation, disagreement adjudication, and completed human validation of all 257 papers are detailed in \autoref{sec:annotation-details}.

\paragraph{Coverage}
The corpus supports a historical synthesis of evaluation concerns and proposed responses, together with analyses of their distribution across research areas and changing prominence over time. We also examine how these concerns are reflected in institutional research checklists. Our coverage remains shaped by the selected sources, search vocabulary, and inclusion criteria: relevant work outside the retrieved venues or using different terminology may have been missed. Accordingly, the patterns reported characterize the reviewed corpus rather than the full evaluation literature.

\paragraph{Access and Intended Use conditions.}The reviewed publications retain their originalcopyright and licensing conditions. ACL's copyrightpolicy distinguishes ACL materials published before2016, licensed under CC BY-NC-SA 3.0, from those
published from 2016 onward, licensed under CC BY 4.0.This policy does not cover third-party materialshosted in the Anthology.\footnote{
\url{https://aclanthology.org/faq/copyright/}}Publications discovered through Semantic Scholar remain subject to the conditions of their original
sources. The corpus and taxonomy are intended to support research on evaluation methodology and historical analysis of the reviewed literature.

\section{PRISMA-ScR Reporting Checklist}
\label{sec:appprisma}

\autoref{tab:prisma-checklist} maps the PRISMA-ScR reporting items \cite{tricco2018prisma} to the corresponding manuscript sections and corpus records, with notes on the information provided.

\paragraph{Selection summary}
As described in \autoref{sec:appsurveymethod}, keyword searches yielded approximately 5,000 candidate records. Human screening against the inclusion criteria retained 194 papers. Following backward citation screening, the final corpus comprised 257 papers: 192 from the ACL Anthology and 65 from the broader
ML collection, spanning 1981--2024. 

\paragraph{Corpus records}
Paper-level information is stored in relevant csv files- one for ACL and one for ML venues. Both files record each paper's title, authors, publication year, source link, topic, research area, summary, and methodological position, alongside the automated annotations, automated adjudication, and final concern tags. Taxonomy construction and the annotation and human-validation procedures are described in \autoref{sec:annotation-details}.

\begin{table*}[t]
\centering
\footnotesize
\setlength{\tabcolsep}{4pt}
\renewcommand{\arraystretch}{1.12}
\begin{tabular}{@{}
    p{0.035\textwidth}
    p{0.235\textwidth}
    p{0.68\textwidth}
@{}}
\toprule
\textbf{\#} &
\textbf{Reporting item} &
\textbf{Location and details} \\
\midrule

\multicolumn{3}{@{}l}{\textit{Title and abstract}} \\[2pt]

1 & Identification as a scoping review &
Identified in the abstract and \S2, rather than the title. \\

2 & Structured summary &
The abstract presents the motivation, approach, and
contribution. Eligibility criteria, search sources,
and charting procedures are described in \S2 and
Apps.~\ref{sec:appsurveymethod}
and \ref{sec:annotation-details}, rather than the abstract. \\

\midrule
\multicolumn{3}{@{}l}{\textit{Introduction}} \\[2pt]

3 & Rationale &
\S1 and \S2; the rationale for the review's scope is also
described in App.~\ref{sec:appsurveymethod}. \\

4 & Objectives &
\S1 and \S2 describe the review's objectives and focus
on methodological concerns in NLP evaluation. \\

\midrule
\multicolumn{3}{@{}l}{\textit{Methods}} \\[2pt]

5 & Protocol and registration &
The review was not pre-registered. The existence and
availability of a separate review protocol are not reported. \\

6 & Eligibility criteria &
App.~\ref{sec:appsurveymethod} gives the publication period
and substantive inclusion and exclusion criteria.
Language and publication-status restrictions are not specified. \\

7 & Information sources &
App.~\ref{sec:appsurveymethod} identifies the bibliographic
sources, coverage period, and citation-based expansion.
The date of the latest search is not reported. \\

8 & Search strategy &
App.~\ref{sec:appsurveymethod} describes keywords and filters.
Complete executable queries, searched fields, and specific
venue filters are not reported. \\

9 & Selection process &
App.~\ref{sec:appsurveymethod} describes human screening,
conservative handling of borderline cases, and backward
citation screening. Screener responsibilities, duplicate
handling, and any separate full-text eligibility stage
are not specified. \\

10 & Data charting process &
\S2 and Apps.~\ref{sec:appsurveymethod}
and \ref{sec:annotation-details} describe manual charting,
LLM annotation, adjudication, and validation of all
257 papers. Whether manual metadata charting was conducted
independently or in duplicate is not specified. \\

11 & Data items &
\S2, Apps.~\ref{sec:appsurveymethod}
and \ref{sec:annotation-details}, and corpus records:
bibliographic fields, contextual annotations, concern
definitions, and tag-assignment rules. \\

12 & Critical appraisal methods &
N/A: no formal appraisal of the included studies' quality
is reported. Human validation in
App.~\ref{sec:annotation-details} concerns annotation
accuracy. \\

13 & Synthesis methods &
\S2 and App.~\ref{sec:annotation-details} describe thematic
synthesis and taxonomy construction; \S7 describes temporal
comparisons and institutional-checklist mapping. \\

\midrule
\multicolumn{3}{@{}l}{\textit{Results}} \\[2pt]

14 & Selection results &
App.~\ref{sec:appsurveymethod} and the selection summary
above report approximately 5,000 candidates, 194 initially
retained papers, and 257 final inclusions. Complete
stage-specific counts and exclusion breakdowns are not reported. \\

15 & Characteristics of sources &
Corpus records provide paper-level metadata and source
links. \S2, \autoref{fig:acl-papers},
\autoref{fig:area-papers}, and
\autoref{tab:collection-dimensions} summarize corpus
characteristics and concern distributions. \\

16 & Critical appraisal results &
N/A: no formal study-quality appraisal results are reported. \\

17 & Results for individual sources &
Corpus records contain each paper's summary, methodological
position, and final concern tags; \S3--6 synthesize the
corresponding arguments and proposed responses. \\

18 & Synthesis of results &
\S3--6 present the concern-based synthesis;
\S7 examines temporal patterns and checklist coverage. \\

\midrule
\multicolumn{3}{@{}l}{\textit{Discussion}} \\[2pt]

19 & Summary of evidence &
\S7 and \S8 discuss the principal findings and their
relevance to the research community. \\

20 & Limitations &
The Limitations section discusses corpus coverage,
interpretive classification, and the scope of the
checklist analysis. \\

21 & Conclusions &
\S8 presents the implications and planned development
of a community-contributed repository; \S7 identifies
opportunities for renewed attention and guidance. \\

\midrule
\multicolumn{3}{@{}l}{\textit{Funding}} \\[2pt]

22 & Funding and funder roles &
Funding of the review and the included papers, and
the role of any review funders, are not reported. \\

\bottomrule
\end{tabular}
\caption{PRISMA-ScR reporting map, with item descriptions
abbreviated. Items 12 and 16 are conditional on conducting
a formal critical appraisal; annotation validation is
reported separately in App.~\ref{sec:annotation-details}.}
\label{tab:prisma-checklist}
\end{table*}

\section{Corpus Annotation Methodology}
\label{sec:annotation-details}

\subsection{Tagset Construction}

For each of the 257 included papers, we conducted a human review and wrote a summary of its evaluation concerns. Through iterative coding and thematic synthesis \cite{thomas2008methods}, we grouped related concerns and refined their scope through discussion among the authors. This process produced 22 concern tags across four dimensions: Data (7), Metrics (7), Hypothesis (6), and Reporting (2). We consolidated their definitions into a shared codebook, following recommendations for explicit annotation schemes \cite{klie-etal-2024-analyzing}. The definitions are provided in \autoref{tab:codebook-data}, \autoref{tab:codebook-metrics}, \autoref{tab:codebook-hypothesis}, and \autoref{tab:codebook-reporting}.

\subsection{Automated Annotation}

\paragraph{Model settings and access.}
We used Google's Gemini 3.8 Flash\footnote{
\url{https://ai.google.dev/gemini-api/docs/models/gemini-3.8-flash}}
and Anthropic's Claude Fable 5.1\footnote{
\url{https://platform.claude.com/docs/en/models/fable-5-1/overview}}
as independent annotators, accessed through GitHub
Copilot subscriptions.\footnote{
\url{https://docs.github.com/en/copilot/reference/ai-models/supported-models}}
We used the settings supplied by GitHub Copilot,
without additional inference-configuration changes.
We did not override sampling or reasoning settings.
We therefore describe the configuration as
service-managed defaults rather than assign fixed
numerical values to parameters that we did not
explicitly configure.
Access was subscription-based, under GitHub Copilot's
plan and usage arrangements.\footnote{
\url{https://docs.github.com/en/copilot/get-started/plans}}

\paragraph{Annotation procedure.}
We used AnnoLLM's explain-then-annotate procedure
\cite{he-etal-2024-annollm}, following research on
LLM-based annotation \cite{tan-etal-2024-large}.
Both annotators used the same concern codebook and
labeled demonstrations with explanations.\footnote{We will release the complete annotation prompts, labeled demonstrations, and associated
instructions after publication.}

\textit{Explain.}
For each labeled demonstration, the LLM received the
task description, concern definitions, paper text,
and established tag set, and was asked to explain
why those tags applied. The resulting explanations
connected the example labels to the codebook and
the paper's methodological concerns.

\textit{Annotate.}
We incorporated the labeled demonstrations and their
explanations into the annotation prompts. Each
annotator independently read the target paper's
extracted text, produced a one- or two-sentence
summary of the evaluation problem and its supporting
evidence, and assigned the most specific eligible
concern tags. The instructions preferred a single
tag and permitted additional tags only for distinct,
substantive concerns, excluding passing mentions.
Annotators could not invent labels.

The human-assigned screening field
\textit{Dimension} restricted eligible tags to the
corresponding taxonomy dimension: Data, Metrics,
Hypothesis, or Reporting. Papers marked
\textit{General} could receive tags across all four
dimensions. Research area imposed no restriction.

\paragraph{Agreement before adjudication.}
We measured agreement between the two annotators
using three complementary measures
(\autoref{tab:annotation-agreement}).
\textit{Exact agreement} required identical
unordered tag sets; differences in generated
summaries were excluded.
\textit{Mean Jaccard similarity} captured partial
overlap by averaging the intersection-to-union
ratio of the tag sets across papers.
\textit{Cohen's kappa} measured chance-corrected
agreement, treating each distinct complete tag set
as a nominal category.

Across all 257 papers, exact agreement was 84.4\%
(217/257), mean Jaccard similarity was 0.857, and
Cohen's $\kappa$ was 0.835.
Exact agreement was 89.6\% (172/192) for ACL papers
and 69.2\% (45/65) for broader ML papers.
These statistics describe agreement between the
initial automated annotations, before adjudication
and human validation; they do not measure annotation
accuracy against an independent gold standard.

\begin{table}[t]
    \centering
    \small
    \begin{tabular}{@{}lr@{}}
        \toprule
        Measure & Value \\
        \midrule
        Exact tag-set agreement & 84.4\% (217/257) \\
        Mean Jaccard similarity & 0.857 \\
        Cohen's $\kappa$ & 0.835 \\
        \bottomrule
    \end{tabular}
    \caption{Agreement between the two LLM annotators
    across all 257 papers, before adjudication and
    human validation.}
    \label{tab:annotation-agreement}
\end{table}

\subsection{Automated Adjudication}

\paragraph{Model settings and access.} We used OpenAI's GPT-6 Astra\footnote{
\url{https://developers.openai.com/api/docs/models/gpt-6-astra}}
as the adjudicator, accessed through GitHub Copilot
subscriptions.\footnote{ \url{https://docs.github.com/en/copilot/reference/ai-models/supported-models}}. As with the initial annotators, we used the settings
supplied by GitHub Copilot without additional inference-configuration changes. We did not override sampling or reasoning settings and describe these
as service-managed defaults. The subscription access described in Appendix D.2 and resource-reporting limitations in Appendix D.5 also apply to adjudication.

\paragraph{Adjudication procedure.}
We adapted the CLEV (Consensus via Lightweight Efficient Voting) process
\cite{badshah-etal-2025-clev} for concern annotation. CLEV obtains judgments from two LLMs and invokes a third when they disagree; in its original binary
question-answering evaluation setting, the third judgment establishes a majority verdict. We followed this selective use of a third judge,
adapting the procedure to adjudicate multi-label tag sets against the source papers. \footnote{We will release the complete adjudication prompts and associated instructions after publication.}

The 217 matching tag sets from the initial annotation stage were provisionally retained, and the remaining 40 papers were routed to GPT-6 Astra. For each disagreement, the adjudicator examined the source paper and both proposed summaries and tag sets against the concern codebook. It could retain either proposal or produce a better-supported eligible tag set, including concerns
omitted by both annotators, with a rationale grounded in identifiable evidence from the paper. The same eligibility restrictions based on the human-assigned \textit{Dimension} field applied.

This adaptation used source-based adjudication of complete tag sets rather than CLEV's binary majority vote. All resulting assignments, including the 217 initially matching tag sets, subsequently underwent human validation as described next.

\subsection{Human Validation}

We adopted the human verification workflow of MEGAnno+ \cite{kim-etal-2024-meganno}. In this workflow, LLMs generate initial labels, and human reviewers inspect the underlying data to confirm or correct them. MEGAnno+ supports selecting annotations for review using labels and metadata, while retaining the original predictions and human corrections. We applied the verification stage to our entire corpus, manually reviewing all 257 papers, including both matching and adjudicated assignments.

For each paper, we checked every proposed tag against the source text and its codebook definition. Tags required substantive engagement with a concern; passing mentions or routine use of an evaluation method were insufficient. We also examined the paper for relevant concerns omitted from the proposed set. Corrections and their reasons were documented, and ambiguous assignments were resolved through discussion among the authors. The resulting tag sets formed the final human-validated corpus.

\subsection{Computation and Software}

We accessed remotely hosted models through GitHub Copilot subscriptions, using service-managed settings without additional configuration (Appendices D.2--D.3).
The final corpus involved two initial annotations per paper for 257 papers and adjudication of 40 disagreements. These counts describe annotation decisions, not total API requests. We do not report a study-specific monetary
cost, model parameter counts, provider-side hardware, GPU hours, or total token usage since these are not available due to our methods of access.

Our Python pipeline uses \texttt{pypdf} for page-wise PDF text extraction, custom standard-library code for exact agreement, mean Jaccard similarity, and Cohen's kappa, and NumPy and Matplotlib for numerical analysis
and figures. The project specifies dependency ranges of \texttt{pypdf} $\geq 6.0, <7$, NumPy $\geq 1.24, <3$, and Matplotlib $\geq 3.7, <4$.

\begin{table*}[t]
\centering
\small
\renewcommand{\arraystretch}{1.15}
\begin{tabular}{@{}p{0.28\textwidth}
    p{\dimexpr0.72\textwidth-2\tabcolsep\relax}@{}}
\toprule
\textbf{Concern} & \textbf{Definition} \\
\midrule

\textbf{Construct Validity} &
Whether evaluation tasks and data measure the intended
capability or concept rather than reward unrelated
behavior. \\
\addlinespace[3pt]

\textbf{Gold Standard Validity} &
Whether reference labels provide defensible ground truth,
accounting for annotation errors, uncertainty, and
legitimate disagreement. \\
\addlinespace[3pt]

\textbf{Artifact Issues} &
Whether systematic cues or biases in evaluation data
allow shortcuts that inflate apparent task competence. \\
\addlinespace[3pt]

\textbf{Contamination Issues} &
Whether exposure to evaluation data during training or
development compromises the independence and credibility
of results. \\
\addlinespace[3pt]

\textbf{Test-Train Splits} &
How data partitioning and held-out sampling affect
estimates of generalization to the intended evaluation
or deployment setting. \\
\addlinespace[3pt]

\textbf{Random vs. Adversarial} &
How randomly sampled and deliberately challenging
evaluation data reveal different aspects of model
performance and robustness. \\
\addlinespace[3pt]

\textbf{Empirical vs. Rational Stance} &
Whether evaluation should rely on naturally occurring
data or theory-driven examples to assess linguistic
performance and competence. \\

\bottomrule
\end{tabular}
\caption{Annotation definitions for data concerns.}
\label{tab:codebook-data}
\end{table*}

% METRICS
\begin{table*}[t]
\centering
\small
\renewcommand{\arraystretch}{1.15}
\begin{tabular}{@{}p{0.28\textwidth}
    p{\dimexpr0.72\textwidth-2\tabcolsep\relax}@{}}
\toprule
\textbf{Concern} & \textbf{Definition} \\
\midrule

\textbf{Downstream Validity} &
Whether metric scores and improvements correspond to
practical usefulness or performance on downstream
tasks. \\
\addlinespace[3pt]

\textbf{Linguistic Validity} &
Whether metrics capture relevant linguistic properties,
such as meaning, syntax, and discourse structure. \\
\addlinespace[3pt]

\textbf{Human Evaluation} &
Whether human judgments provide valid and reliable
evaluation, considering evaluator selection,
instructions, rating methods, and agreement. \\
\addlinespace[3pt]

\textbf{Data Sensitivity} &
How metric behavior and interpretation change with
data characteristics such as composition, size,
length, or language. \\
\addlinespace[3pt]

\textbf{Task Sensitivity} &
Whether a metric captures the properties that matter
for a particular task and remains appropriate when
used across tasks. \\
\addlinespace[3pt]

\textbf{Model Sensitivity} &
How metric behavior depends on the models being
evaluated, their training or compression, or the
model used as an automated evaluator. \\
\addlinespace[3pt]

\textbf{Metric Standardization} &
How metrics should be selected, combined, and used
consistently to support meaningful comparisons
across studies. \\

\bottomrule
\end{tabular}
\caption{Annotation definitions for metric concerns.}
\label{tab:codebook-metrics}
\end{table*}

% HYPOTHESIS
\begin{table*}[t]
\centering
\small
\renewcommand{\arraystretch}{1.15}
\begin{tabular}{@{}p{0.28\textwidth}
    p{\dimexpr0.72\textwidth-2\tabcolsep\relax}@{}}
\toprule
\textbf{Concern} & \textbf{Definition} \\
\midrule

\textbf{Relevance of Hypothesis} &
Whether evaluation tests explicit, meaningful hypotheses
about system capabilities, requirements, or usefulness. \\
\addlinespace[3pt]

\textbf{Task-Specificity of Hypothesis} &
Whether the task used for evaluation justifies the
hypothesis or broader claim being made about
model quality. \\
\addlinespace[3pt]

\textbf{Reliability of Statistical Tests} &
Whether statistical tests and their assumptions
support valid inferences about observed performance
differences. \\
\addlinespace[3pt]

\textbf{Parameters of Statistical Tests} &
How significance thresholds, sample sizes, statistical
power, and effect sizes should be chosen or estimated
for meaningful inference. \\
\addlinespace[3pt]

\textbf{Experimental Design} &
How baselines, controls, ablations, and comparison
settings support fair comparisons and attribution
of performance differences. \\
\addlinespace[3pt]

\textbf{Leaderboard Practices} &
How the design, use, and interpretation of model
rankings shape comparisons, research incentives,
and claims of progress. \\

\bottomrule
\end{tabular}
\caption{Annotation definitions for hypothesis concerns.}
\label{tab:codebook-hypothesis}
\end{table*}

% REPORTING
\begin{table*}[t]
\centering
\small
\renewcommand{\arraystretch}{1.15}
\begin{tabular}{@{}p{0.28\textwidth}
    p{\dimexpr0.72\textwidth-2\tabcolsep\relax}@{}}
\toprule
\textbf{Concern} & \textbf{Definition} \\
\midrule

\textbf{Transparency Concerns} &
Whether evaluation reports disclose the assumptions,
methods, resources, and limitations needed to
interpret results. \\
\addlinespace[3pt]

\textbf{Reproducibility Concerns} &
Whether evaluation results can be reproduced or
replicated, and what methodological or resource
barriers prevent this. \\

\bottomrule
\end{tabular}
\caption{Annotation definitions for reporting concerns.}
\label{tab:codebook-reporting}
\end{table*}

\section{Evaluation Checklist}
\label{sec:appchecklist}

The checklist in \autoref{fig:eval-checklist} translates the 22 taxonomy
concerns into prompts for evaluation design, interpretation, and reporting.
Each prompt retains the name of its corresponding concern and has a short
identifier: D1--D7 for Data, M1--M7 for Metrics, H1--H6 for Hypothesis,
and R1--R2 for Reporting. The definitions in
Tables~\ref{tab:codebook-data}--\ref{tab:codebook-reporting} specify their
scope, and the main survey discusses the supporting literature.

The checklist can be used when planning an evaluation and revisited when
interpreting and reporting results. For each item, record a status
(\textit{addressed}, \textit{open}, or \textit{not applicable}) together
with a brief justification and, where available, a reference to supporting
evidence. An addressed item may document a justified design choice or an
acknowledged limitation. Open items identify decisions or checks that
remain unresolved; non-applicable items require a reason. The identifiers
allow these records to be linked to individual concerns. The checklist
has not been validated as a scoring instrument, and we do not aggregate
responses into an evaluation-quality score.

The checklist complements task- and modality-specific guidance for human
evaluation and NLG
\cite{van-der-lee-etal-2019-best,belz-etal-2020-disentangling,howcroft-etal-2020-twenty}
by bringing together concerns across the evaluation pipeline.

\begin{figure*}[t]
\centering
\begingroup
\definecolor{checkBlue}{HTML}{244A70}
\definecolor{checkPaleBlue}{HTML}{EAF0F6}
\definecolor{checkGray}{HTML}{66717D}
\definecolor{checkBorder}{HTML}{C6CDD5}
\begin{tcolorbox}[
  enhanced,
  title={Evaluation Checklist},
  colback=white, colframe=checkBorder,
  colbacktitle=checkBlue, coltitle=white,
  fonttitle=\sffamily\bfseries,
  boxrule=0.6pt, arc=2pt,
  left=8pt, right=8pt, top=5pt, bottom=5pt,
  width=\textwidth,
]
\small\sffamily
{\color{checkGray}\raggedright
For each item: \textbf{addressed / open / not applicable}.
Record a justification and an evidence reference where available.\par}

\tcbsubtitle[
  colback=checkPaleBlue, colframe=checkBorder,
  coltitle=checkBlue, before skip=4pt, after skip=4pt,
]{\color{checkBlue}\textbf{Data}\hfill{\normalfont\sffamily\small Evidence and sampling}}
\begin{minipage}[t]{0.48\linewidth}
\raggedright
\begin{itemize}[leftmargin=1.9em, labelsep=0.45em,
  itemsep=4pt, topsep=2pt, parsep=0pt]
  \item[\textcolor{checkBlue}{\textbf{D1}}]
  \textbf{Construct Validity.} What evidence shows that the task and data measure the intended capability?
  \item[\textcolor{checkBlue}{\textbf{D2}}]
  \textbf{Gold Standard Validity.} How are reference-label errors, uncertainty, and legitimate disagreement assessed and handled?
  \item[\textcolor{checkBlue}{\textbf{D3}}]
  \textbf{Artifact Issues.} Which dataset cues could enable shortcuts, and how have their effects been checked?
  \item[\textcolor{checkBlue}{\textbf{D4}}]
  \textbf{Contamination Issues.} How was prior exposure to evaluation data assessed, and what mitigation or uncertainty remains?
\end{itemize}
\end{minipage}%
\hfill
\begin{minipage}[t]{0.48\linewidth}
\raggedright
\begin{itemize}[leftmargin=1.9em, labelsep=0.45em,
  itemsep=4pt, topsep=2pt, parsep=0pt]
  \item[\textcolor{checkBlue}{\textbf{D5}}]
  \textbf{Test-Train Splits.} How do partitioning and held-out sampling support the intended generalization claim?
  \item[\textcolor{checkBlue}{\textbf{D6}}]
  \textbf{Random vs. Adversarial.} Why are randomly sampled examples, deliberately challenging examples, or both appropriate for the evaluation goal?
  \item[\textcolor{checkBlue}{\textbf{D7}}]
  \textbf{Empirical vs. Rational Stance.} Why are naturally occurring data, theory-driven examples, or both appropriate for the claim?
\end{itemize}
\end{minipage}%

\tcbsubtitle[
  colback=checkPaleBlue, colframe=checkBorder,
  coltitle=checkBlue, before skip=4pt, after skip=4pt,
]{\color{checkBlue}\textbf{Metrics}\hfill{\normalfont\sffamily\small Measurement and interpretation}}
\begin{minipage}[t]{0.48\linewidth}
\raggedright
\begin{itemize}[leftmargin=1.9em, labelsep=0.45em,
  itemsep=4pt, topsep=2pt, parsep=0pt]
  \item[\textcolor{checkBlue}{\textbf{M1}}]
  \textbf{Downstream Validity.} What evidence connects metric improvements to practical usefulness or downstream performance?
  \item[\textcolor{checkBlue}{\textbf{M2}}]
  \textbf{Linguistic Validity.} Which relevant linguistic properties does the metric capture, and which does it miss?
  \item[\textcolor{checkBlue}{\textbf{M3}}]
  \textbf{Human Evaluation.} How are evaluator selection, instructions, rating methods, and judgment reliability justified or assessed?
  \item[\textcolor{checkBlue}{\textbf{M4}}]
  \textbf{Data Sensitivity.} How does metric behavior change with relevant differences in data composition, size, length, or language?
\end{itemize}
\end{minipage}%
\hfill
\begin{minipage}[t]{0.48\linewidth}
\raggedright
\begin{itemize}[leftmargin=1.9em, labelsep=0.45em,
  itemsep=4pt, topsep=2pt, parsep=0pt]
  \item[\textcolor{checkBlue}{\textbf{M5}}]
  \textbf{Task Sensitivity.} Why is the metric appropriate for this task, including any transfer from another task?
  \item[\textcolor{checkBlue}{\textbf{M6}}]
  \textbf{Model Sensitivity.} Could the metric favor particular model types or reflect automated-evaluator biases, and how was this checked?
  \item[\textcolor{checkBlue}{\textbf{M7}}]
  \textbf{Metric Standardization.} How are metric selection, combination, and implementation justified and specified for comparison across studies?
\end{itemize}
\end{minipage}%

\tcbsubtitle[
  colback=checkPaleBlue, colframe=checkBorder,
  coltitle=checkBlue, before skip=4pt, after skip=4pt,
]{\color{checkBlue}\textbf{Hypothesis}\hfill{\normalfont\sffamily\small Claims and comparisons}}
\begin{minipage}[t]{0.48\linewidth}
\raggedright
\begin{itemize}[leftmargin=1.9em, labelsep=0.45em,
  itemsep=4pt, topsep=2pt, parsep=0pt]
  \item[\textcolor{checkBlue}{\textbf{H1}}]
  \textbf{Relevance of Hypothesis.} What explicit hypothesis is tested, and why does it matter for system capabilities or usefulness?
  \item[\textcolor{checkBlue}{\textbf{H2}}]
  \textbf{Task-Specificity of Hypothesis.} How does performance on the chosen task support the hypothesis and the scope of the claim?
  \item[\textcolor{checkBlue}{\textbf{H3}}]
  \textbf{Reliability of Statistical Tests.} Why are the tests and their assumptions appropriate for the data, dependencies, and comparisons?
\end{itemize}
\end{minipage}%
\hfill
\begin{minipage}[t]{0.48\linewidth}
\raggedright
\begin{itemize}[leftmargin=1.9em, labelsep=0.45em,
  itemsep=4pt, topsep=2pt, parsep=0pt]
  \item[\textcolor{checkBlue}{\textbf{H4}}]
  \textbf{Parameters of Statistical Tests.} How are significance thresholds, sample sizes, power, and effect sizes justified or assessed?
  \item[\textcolor{checkBlue}{\textbf{H5}}]
  \textbf{Experimental Design.} How do baselines, controls, ablations, and comparison settings support fair comparison and attribution?
  \item[\textcolor{checkBlue}{\textbf{H6}}]
  \textbf{Leaderboard Practices.} How robust are ranking claims to uncertainty and evaluation choices, and what do the rankings justify?
\end{itemize}
\end{minipage}%

\tcbsubtitle[
  colback=checkPaleBlue, colframe=checkBorder,
  coltitle=checkBlue, before skip=4pt, after skip=4pt,
]{\color{checkBlue}\textbf{Reporting}\hfill{\normalfont\sffamily\small Documentation and reuse}}
\begin{minipage}[t]{0.48\linewidth}
\raggedright
\begin{itemize}[leftmargin=1.9em, labelsep=0.45em,
  itemsep=4pt, topsep=2pt, parsep=0pt]
  \item[\textcolor{checkBlue}{\textbf{R1}}]
  \textbf{Transparency Concerns.} Where are the assumptions, methods, resources, and limitations needed to interpret the results documented?
\end{itemize}
\end{minipage}%
\hfill
\begin{minipage}[t]{0.48\linewidth}
\raggedright
\begin{itemize}[leftmargin=1.9em, labelsep=0.45em,
  itemsep=4pt, topsep=2pt, parsep=0pt]
  \item[\textcolor{checkBlue}{\textbf{R2}}]
  \textbf{Reproducibility Concerns.} What procedures, code, data, and configurations support reproduction or replication, and which access barriers remain?
\end{itemize}
\end{minipage}%

\end{tcolorbox}
\endgroup
\caption{Evaluation checklist covering all 22 taxonomy concerns.
Item identifiers link each prompt to a recorded status, justification,
and supporting evidence. Open items and acknowledged limitations can
inform further evaluation or qualify the resulting claims.}
\label{fig:eval-checklist}
\end{figure*}

\end{document}